\documentclass{article}

\usepackage[preprint]{neurips_2026}

\usepackage[utf8]{inputenc}
\usepackage[T1]{fontenc}
\usepackage{url}
\usepackage{booktabs}
\usepackage{amsfonts}
\usepackage{amsmath}
\usepackage{nicefrac}
\usepackage{microtype}
\usepackage{xcolor}
\usepackage{graphicx}
\usepackage{hyperref}

\title{Scaling Laws for Classical Machine Learning on Tabular Data: An Empirical Benchmark Study}

\author{%
  Kaihua Ding\\
  University of Pennsylvania\\
  \texttt{dkaihua@upenn.edu}%
}

\begin{document}

\maketitle

\begin{abstract}
Prior classical-ML learning-curve work fits power laws to tree, linear,
and kernel models on tabular data \citep{perlich2003tree,
mukherjee2003estimating, figueroa2012predicting, domhan2015speeding},
but at small scale: typically one curve, one team, a handful of cells.
We present a distributed benchmark study of classical-ML scaling laws: a
graduate cohort of 127 independent replicators each ran a fixed protocol
on 3 assigned datasets, drawn from 18 tabular classification and
regression datasets and 6 model families (Boosting, Random Forest, SVM,
Linear/Logistic, Ridge, Lasso), yielding 11{,}536 training runs and
1{,}648 fitted power-law curves of the form
$\mathrm{error}(N)=aN^{-b}+c$. Because each replicator was assigned a
different subset of datasets, this is a partial rather than a full
replication: our focus is the scaling exponents and whether they are
shared across a model family, not exact reproduction of a single result.
Three findings.
\textbf{(1)~Power laws fit:} $R^2>0.8$ on 77.7\% of cells, with tree
ensembles dominating at full data (Boosting 50\% of datasets,
RandomForest 33\%; linear models underperform on classification).
\textbf{(2)~Approximate shared exponents within a model family:} for 5
of 6 families, a single family-level exponent predicts each family's
cross-dataset curves nearly as well as per-dataset exponents ($R^2$
gap $<0.011$), though AIC favors the unconstrained fit and curve
collapse is partial (32--58\% of points within $\pm 0.5$\,dex). We
frame this as approximate predictive compressibility, not
dataset-independent universality; Lasso fails outright (negative
control) and Ridge is fragile under leave-one-dataset-out.
\textbf{(3)~Replicator-implementation variance:} with the random seed
fixed, independent re-implementations of the same protocol still differ
by a mean coefficient of variation $\mathrm{CV}(b)=0.144$ on the fitted
exponent $b$---here CV is the across-replicator standard deviation
divided by the mean, so $0.144$ means the fitted exponent moves by about
$14\%$ from one compliant re-implementation to the next. This is not seed
variance, but the spread induced by unconstrained parts of the protocol
(preprocessing, encoding, missing-value handling). We release the
aggregated curves, per-cell fits, and a practical data-requirement table
for $N^\star$---the training-set size a model needs to reach a target
test error, here set to $0.15$ (e.g.\ $1-\mathrm{AUROC}=0.15$, i.e.\
AUROC $\approx 0.85$, a practical useful-accuracy threshold for these
tasks).

\end{abstract}

\section{Introduction}
\label{sec:introduction}

Empirical scaling laws have become central to how large neural networks are
designed, trained, and budgeted: predictable power-law decay of test loss with
model size, dataset size, and compute now informs decisions worth tens of
millions of dollars per training run \citep{kaplan2020scaling,
hoffmann2022training, hestness2017deep, rosenfeld2020constructive}. Yet the
overwhelming majority of applied machine learning in industry is not large
neural networks on text or images. It is gradient-boosted trees, random
forests, and linear models on tabular business data---churn prediction,
default risk, sales forecasting, ad-click prediction, and causal market segmentation \citep{ding2024iterative}. These are precisely the
settings where tree ensembles continue to outperform deep learning
\citep{grinsztajn2022why, shwartzziv2022tabular}. While a classical-ML
learning-curve literature exists \citep{perlich2003tree,
mukherjee2003estimating, figueroa2012predicting, domhan2015speeding}, no
prior study to our knowledge characterizes how their generalization error
scales with training-set size at the cross-replicator scale that Kaplan
et al.\ achieved for transformers within a single research team.

Our contribution is not the idea of fitting a power law to a tree- or
linear-model learning curve but \emph{scale}: we run a single fixed
protocol 127 times, once per member of a graduate ML cohort, with each
member independently assigned a different subset of 3 of the 18 datasets.
The 127 runs are therefore spread across the full dataset suite rather
than being 127 repeats of one setting, yielding 11{,}536 model fits and
1{,}648 successful scaling curves. This lets us measure both the
exponents and their variance across nominally compliant
re-implementations.

Our contributions are:
\begin{itemize}
  \item \textbf{Approximate shared exponents within a model family}:
        We separate three levels of claim. (a)~\emph{Fit}: the
        additive power law describes individual curves well (77.7\%
        with $R^2>0.8$). (b)~\emph{Compression}: a single
        family-level $b_m$ replaces per-dataset $b_{m,d}$ with small
        predictive loss ($R^2$ gap $<0.011$ on 5 of 6 families;
        held-out leave-one-dataset-out (LODO) gap $\le 0.012$ for Boosting, RandomForest,
        LinearModel). (c)~\emph{Universality} ($b_{m,d}=b_m$):
        \emph{not supported}---AIC favors per-dataset fits, curve
        collapse is partial, and a permutation null shows the
        in-sample gap lacks specificity. Lasso fails outright as a
        negative control; Ridge is fragile (LODO CV $0.96$). The
        contribution is the compression result, not universality.
  \item \textbf{Empirical benchmark}: 18 datasets $\times$ 6 model
        families $\times$ 7 nested training fractions, 11{,}536 runs
        aggregated from a graduate cohort of 127 replicators; tree ensembles win at full data
        (Boosting 50\%, RandomForest 33\%).
  \item \textbf{Replicator-implementation variance}: With the random
        seed fixed, $\mathrm{CV}(b)=0.144$ across replicators measures
        unconstrained-protocol spread (preprocessing, encoding,
        missing-value handling), not seed variance.
  \item \textbf{Practical data-requirement curves}: pooled-fit
        $N^\star$ to reach $0.15$ target error, with cells where the
        floor $c\ge 0.15$ flagged as unreachable for that model.
\end{itemize}

\section{Related Work}
\label{sec:related}

\paragraph{Neural scaling laws.}
A line of work beginning with \citet{hestness2017deep} and crystallized by
\citet{kaplan2020scaling} established that neural language-model test loss
follows a clean power law in model size, dataset size, and compute. The
Chinchilla paper \citep{hoffmann2022training} refined the joint
parameter--data tradeoff and showed that prior large models were systematically
undertrained. \citet{rosenfeld2020constructive} extended the formalism to
image classification and proposed the additive-irreducible-error
parameterization $aN^{-b}+c$ that we adopt directly. These results are all on
neural models---transformers, ResNets, recurrent nets---and at moderately
large data scales. Whether the same functional form applies to gradient-boosted
trees on a few thousand tabular rows has, until now, not been settled
empirically at scale.

\paragraph{Classical-ML learning curves.}
Fitting power laws to classical-ML learning curves is itself a long-studied
topic that predates the neural-scaling-laws literature. \citet{perlich2003tree}
fit learning curves for tree induction versus logistic regression on a panel
of UCI-style datasets. \citet{mukherjee2003estimating} use learning-curve
extrapolation to estimate sample-size requirements for DNA microarray
classification. \citet{figueroa2012predicting} fit and invert learning curves
to predict the sample size required to reach a target classification
performance. \citet{domhan2015speeding} extrapolate learning curves to speed
up hyperparameter optimization. The broader principle---treating error as a
smooth, extrapolable function of a resource budget and using the resulting
estimate to plan how much budget to spend---is shared with output-based error
estimation in numerical analysis, where an adjoint-weighted error estimate
directs where to add mesh resolution to reduce a target output's error most
efficiently \citep{ding2016adjoint}; the fitted $aN^{-b}+c$ plays the analogous
role for labeled data, telling a practitioner how many additional rows buy a
given error reduction. We do not claim novelty in the idea of
fitting an $aN^{-b}+c$ law to classical models on tabular data; what is
new is the \emph{scale} and the question that scale lets us answer. To our
knowledge this is the largest classical-ML learning-curve study to
date---127 independent replicators $\times$ 18 datasets $\times$ 6 model
families, 1{,}648 fitted curves, orders of magnitude beyond the
single-curve, single-team efforts above---and the first large enough to
test directly whether a \emph{single scaling exponent is shared} across a
model family's datasets, rather than fitting one curve at a time. That
question, not the curve-fitting itself, is the point: prior work
establishes that classical learning curves are power laws; we ask whether
those power laws have a common exponent, and how far a reported exponent
can be trusted. On top of this we contribute a \emph{protocol-level
reproducibility lens} (quantifying how far a fitted exponent drifts across
nominally compliant re-implementations) and a \emph{public release} of the
aggregated curves, per-cell pooled fits, and data-requirement tables as a
benchmark resource.

\paragraph{Tabular ML benchmarking.}
A parallel literature has shown repeatedly that on most tabular tasks,
gradient boosting and random forests still beat deep architectures
\citep{grinsztajn2022why, shwartzziv2022tabular, chen2016xgboost,
breiman2001random}. The standard methodology in that literature, however, is
to evaluate models at \emph{full data}, not to trace error as a function of
$N$. Our work complements this by adding the sample-size axis: the
ranking-at-full-data picture is enriched by knowing how quickly each family
approaches its irreducible-error floor.

\paragraph{Reproducibility and replication.}
\citet{pineau2021improving} document the reproducibility crisis in machine
learning and report on the NeurIPS Reproducibility Program.
\citet{bouthillier2019unreproducible} argue that single-run reports
systematically understate uncertainty: variance from seeds, data splits, and
implementation details is real and large. Quantifying reliability is harder
still when no gold-standard label is available: \citet{ding2025variance}
address this with a variance-bounded evaluation framework that scores a system
by its expected success across plausible interpretations, penalized by
variance. Our replicator-spread estimate is a scaling-law analogue---it treats
the variance of a reported exponent across nominally compliant
re-implementations as a first-class quantity rather than a nuisance. Our
127-replicator design is, in
effect, a controlled multi-replicator experiment whose primary product---in
addition to the scaling laws themselves---is a quantitative estimate of how
much a single research group's reported scaling exponent should be trusted.

\section{Methodology}
\label{sec:methodology}

\subsection{Datasets}
\label{sec:datasets}

We curated 18 publicly available tabular datasets covering business-relevant
classification and regression tasks, drawn primarily from the UCI Machine
Learning Repository \citep{dua2019uci} and other public repositories,
summarized in \autoref{tab:datasets}. Ten
are classification (binary, with class imbalance ranging from 54\%/46\% to
88\%/12\%), and eight are regression. Sample sizes range from 303
(\texttt{heart\_disease} \citep{dua2019uci}) to 53{,}940
(\texttt{diamonds} \citep{wickham2016ggplot2}). Feature counts range
from 4 to 74, mixing numeric and categorical columns. All datasets are widely
used in undergraduate and applied-ML courses, chosen because they admit
straightforward preprocessing (one-hot encoding for categoricals, median
imputation for the small number of missing values) and because the tasks are
well-defined.

\begin{table}[t]
\caption{The 18 curated datasets used in this study.}
\label{tab:datasets}
\centering
\small
\begin{tabular}{lrrll}
\toprule
Dataset & Rows & Features & Task & Domain \\
\midrule
adult\_income         & 45{,}222 & 14 & classification & Census / income \\
bank\_marketing       & 45{,}211 & 16 & classification & Marketing \\
credit\_card\_default & 30{,}000 & 23 & classification & Credit risk \\
diabetes\_pima        &      768 &  8 & classification & Healthcare \\
employee\_attrition   &  1{,}470 & 30 & classification & HR \\
german\_credit        &  1{,}000 & 20 & classification & Credit risk \\
heart\_disease        &      303 & 13 & classification & Healthcare \\
online\_shoppers      & 12{,}330 & 17 & classification & E-commerce \\
telco\_churn          &  7{,}043 & 19 & classification & Telecom \\
wine\_quality         &  6{,}497 & 12 & classification & Quality control \\
abalone               &  4{,}177 &  8 & regression     & Biology \\
ames\_housing         &  1{,}460 & 74 & regression     & Real estate \\
auto\_mpg             &      398 &  7 & regression     & Automotive \\
bike\_sharing         & 17{,}379 & 12 & regression     & Mobility \\
california\_housing   & 20{,}640 &  8 & regression     & Real estate \\
concrete\_strength    &  1{,}030 &  8 & regression     & Materials \\
diamonds              & 53{,}940 &  9 & regression     & Pricing \\
insurance\_charges    &  1{,}338 &  6 & regression     & Insurance \\
\bottomrule
\end{tabular}
\end{table}

\subsection{Models and Protocol}
\label{sec:protocol}

We evaluate six classical model families, all from \texttt{scikit-learn} with
fixed default-style hyperparameters: (i)~\textbf{Boosting}
(\texttt{GradientBoostingClassifier} / \texttt{Regressor});
(ii)~\textbf{Random Forest}; (iii)~\textbf{SVM} (RBF kernel);
(iv)~\textbf{Linear / Logistic} regression; (v)~\textbf{Ridge} regression
(regression only); (vi)~\textbf{Lasso} regression (regression only).
Hyperparameters are fixed across all $N$---no per-$N$ retuning---to keep
results comparable across replicators and to isolate the data-size effect.

For each (dataset, model) pair, each replicator performed a single 80/20 train/test
split with seed 42, then trained on \emph{nested subsets} of the train fold at
the seven fractions $\{0.01,\,0.05,\,0.10,\,0.25,\,0.50,\,0.75,\,1.00\}$ of
the 80\%-train pool. Subsets are nested (smaller subsets are subsets of larger
ones) to reduce sampling noise across fractions. The same held-out 20\% test
set is used to evaluate every fraction.

For classification we report $1-\mathrm{AUROC}$ as the error metric; for
regression we report $\mathrm{RMSE}/\mathrm{std}(y_{\text{test}})$, i.e.\ RMSE
normalized by the test-set standard deviation of the target. To bound SVM
training time, classification SVMs were capped at a 10{,}000-row subsample of
the training fraction.

\subsection{Power-Law Fitting}
\label{sec:fitting}

For each (replicator, dataset, model) triple we fit a three-parameter power law,
\begin{equation}
  \mathrm{error}(N) \;=\; a\, N^{-b} \;+\; c,
  \qquad a,b\ge 0,
  \label{eq:powerlaw}
\end{equation}
to the seven $(N,\mathrm{error})$ points using non-linear least squares.
Bounds on the irreducible-error floor are $c\in[0,1]$ for classification
(since $1-\mathrm{AUROC}\in[0,1]$) and $c\in[0,5]$ for regression. The
exponent $b$ is bounded to $[0,3]$. Fits with $b$ at the upper bound or with
$a$ collapsing to zero are flagged but retained when reporting distributions.
We use the threshold $R^2>0.7$ when reporting per-model exponent statistics
(to exclude obviously failed fits without overpruning), and $R^2>0.8$ when
counting ``successful'' fits in headline statements.

\subsection{Cohort-Scale Aggregation}
\label{sec:aggregation}

The protocol of \autoref{sec:protocol} was executed independently by a
graduate ML cohort of 127 replicators. Each cohort member was assigned three
datasets, spread across classification and regression. Per-(dataset, model)
coverage ranges from 11 to 30 replicators depending on the random assignment
(with SVM coverage slightly lower because a few replicators timed out on the
largest classification datasets). In total this produced 11{,}536 individual
model training runs and 1{,}648 replicator-level power-law fits that converged
(per-replicator fit convergence: 100\%; pooled-method failures: 0/80).

We aggregate per-replicator fits to a population-level summary in two ways. The
\emph{pooled} approach concatenates all $(N, \mathrm{error})$ points from all
replicators for a given (dataset, model) and fits a single power law to the
union; this is what we report for the data-requirement curves. The
\emph{per-fraction-mean} approach first averages error across replicators at
each fraction and fits a power law to those seven means; we use this for
robustness checks. Cross-replicator dispersion is reported as the
\emph{coefficient of variation} $\mathrm{CV}(b) = \sigma_b/\mu_b$---the
across-replicator standard deviation of the fitted value divided by its mean
(a unitless measure of relative spread; not cross-validation and not a raw
variance)---of the scaling-law exponent $b$ of \autoref{eq:powerlaw}, taken
across replicators within a (dataset, model) cell that has at least 5
replicators.

Because each replicator was assigned only 3 of 18 datasets, the per-cell
$\mathrm{CV}(b)$ is computed over a different subset of replicators for
each (dataset, model) cell, so the design is partially blocked rather
than fully crossed. We therefore treat the resulting variance as
observational and the headline $\mathrm{CV}(b)=0.144$ as a marginal
average over heterogeneous cells.

\subsection{Cross-Dataset Combination}
\label{sec:crosscomb}

To test whether a single exponent describes a model family across
datasets, we combine per-cell evidence three ways. (i)~\emph{Per-cell
pooled fit}: \autoref{eq:powerlaw} fit once per (dataset, model) on the
union of replicator points. (ii)~\emph{Shared-exponent joint fit}: for each
family we fit $\mathrm{error}_{d,i}(N) = a_d N^{-b_{\text{model}}} + c_d$
jointly across its datasets, sharing $b_{\text{model}}$ while leaving
$(a_d, c_d)$ free per dataset (\texttt{scipy.optimize.least\_squares}
with the bounds of \autoref{sec:fitting}; soft-$L_1$ loss for Lasso),
compared to an unconstrained joint fit (free $b_d$) by $R^2$ and AIC
on the same point cloud. (iii)~\emph{Random-effects meta-analysis}:
within each cell with $\ge 5$ replicators we pool $\hat b$ via
DerSimonian--Laird, using $(\hat\sigma_b/\sqrt{n})^2$ as within-study
variance, and collapse per-replicator curves onto a normalised axis
$e_{\text{norm}}=(\mathrm{error}-c)/a$ vs.\ $N$.

Upon acceptance we will release the aggregated $(N,\mathrm{error})$ CSVs
for all 11{,}536 runs, per-cell pooled fits, the data-requirement table,
the analysis scripts, and the protocol skeleton distributed to each replicator at
\url{https://github.com/dingkaihua/classical_scaling}.

\section{Results}
\label{sec:results}

\subsection{Power-Law Fits}
\label{sec:fits}

Across all 1{,}648 converged per-replicator fits, 77.7\% achieve $R^2>0.8$ and
74.5\% achieve $R^2>0.9$, supporting the claim that the additive power law
of \autoref{eq:powerlaw} is a useful description of how classical-ML test
error scales with $N$ on tabular data. Fit quality is, however, very uneven
across model families (\autoref{tab:r2}). Tree ensembles, SVM, and
Linear/Logistic regression fit cleanly: 87.9--94.8\% of their per-replicator
fits exceed $R^2>0.8$. Ridge and especially Lasso fit poorly, with only
47.5\% and 23.8\% respectively above that threshold.

\begin{table}[ht]
\caption{$R^2$ distribution per model family (all fits).}
\label{tab:r2}
\centering
\small
\begin{tabular}{lrrrrr}
\toprule
Model & mean & median & q25 & q75 & \%fits $R^2>0.8$ \\
\midrule
Boosting     & 0.954 & 0.985 & 0.949 & 0.999 & 92.4\% \\
Lasso        & 0.546 & 0.649 & 0.443 & 0.703 & 23.8\% \\
LinearModel  & 0.910 & 0.967 & 0.916 & 0.997 & 90.4\% \\
RandomForest & 0.947 & 0.971 & 0.956 & 0.991 & 87.9\% \\
Ridge        & 0.658 & 0.745 & 0.324 & 0.968 & 47.5\% \\
SVM          & 0.961 & 0.978 & 0.952 & 0.997 & 94.8\% \\
\bottomrule
\end{tabular}
\end{table}

The poor Lasso and Ridge fits are mechanistic, not noise. Lasso shrinks all
coefficients to zero in the small-$N$ regime under fixed regularization
strength, producing a flat ``predict-the-mean'' error that is invariant to
$N$ and incompatible with monotone power-law decay. Ridge shows a related
pathology on \texttt{california\_housing}, where the held-out RMSE briefly
\emph{rises} with $N$ at intermediate fractions before settling, which the
power-law form (\autoref{eq:powerlaw}) cannot represent. We flag these
mechanisms rather than discarding the cells: they are real findings about
the limits of the power-law description for regularized linear models with
fixed hyperparameters. The R\textsuperscript{2} heatmap in
\autoref{fig:r2} makes the pattern visible at a glance.

\subsection{Scaling Exponents and Irreducible Error}
\label{sec:exponents}

\begin{figure}[ht]
\centering
\includegraphics[width=0.7\linewidth]{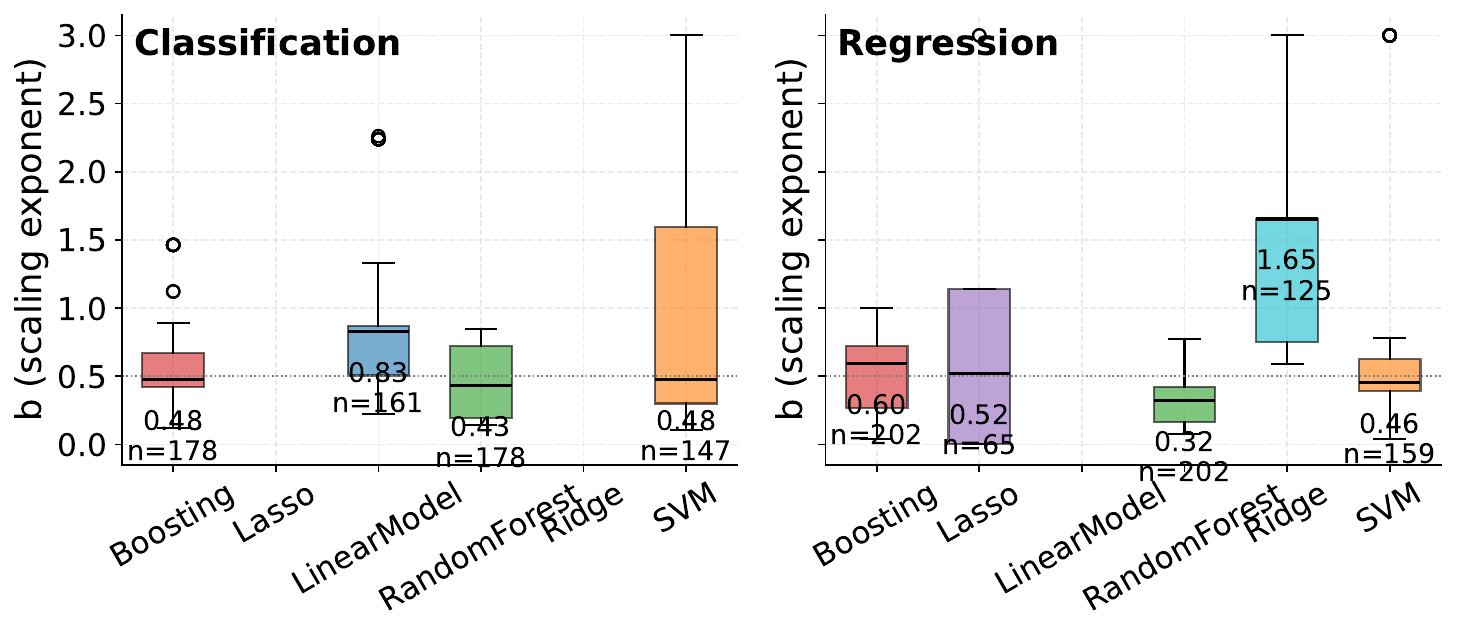}
\caption{Boxplots of fitted scaling exponent $b$ per model, split by task
type. Tree ensembles cluster near $b\approx0.3$--$0.5$; linear models on
classification show systematically larger but noisier exponents.}
\label{fig:b}
\end{figure}

The scaling exponent $b$ varies systematically across model families
(\autoref{fig:b} and \autoref{tab:b}). Random Forest has the smallest median
exponent ($\tilde b=0.325$): its error decays with $N$ but slowly, consistent
with the high-variance averaging behavior of bagged trees. Boosting has a
moderate exponent ($\tilde b=0.480$). Linear/Logistic regression has the
largest stable exponent ($\tilde b=0.830$) on classification, reflecting the
fact that low-capacity models gain more from each additional training point
relative to their already-high error. Ridge has an even larger median exponent
($\tilde b=1.654$) but with very wide spread and poor fit quality; we view
it as a fitting artifact rather than a meaningful exponent.

The median exponent across all good fits is $0.495$. Of classification fits,
47.7\% have $b>0.5$ and 76.5\% have $b>0.3$. Of regression fits, 51.9\% have
$b>0.5$. Consistent with this, classification exponents are slightly higher
than regression exponents within each family that does both: Boosting
($0.575$ clf vs.\ $0.508$ reg), RandomForest ($0.426$ vs.\ $0.325$), and
SVM ($1.023$ vs.\ $0.745$).

\begin{table}[ht]
\caption{Distributions of $b$ and $c$ per model (per-replicator fits with
$R^2>0.7$). The $n$ column reflects only fits passing the $R^2>0.7$
filter; Lasso ($n{=}65$) and Ridge ($n{=}125$) have substantially smaller
$n$ than Boosting/RandomForest/SVM ($n{=}306$--$380$) because their
power-law fits are noisier (cf.\ the $R^2$ heatmap in \autoref{fig:r2}).
The reported $b$ and $c$ for Lasso and Ridge are therefore conditional on
the rare cells where the additive power-law form even applies, not on
the full set of convergent fits.}
\label{tab:b}
\centering
\small
\begin{tabular}{lrrrrr|rrr}
\toprule
& \multicolumn{5}{c|}{$b$ (exponent)} & \multicolumn{3}{c}{$c$ (floor)} \\
Model & mean & median & q25 & q75 & $n$ & mean & median & $n$ \\
\midrule
Boosting     & 0.540 & 0.480 & 0.348 & 0.724 & 380 & 0.213 & 0.208 & 380 \\
Lasso        & 0.634 & 0.522 & 0.001 & 1.142 &  65 & 0.773 & 0.999 &  65 \\
LinearModel  & 0.856 & 0.830 & 0.506 & 0.867 & 161 & 0.139 & 0.139 & 161 \\
RandomForest & 0.372 & 0.325 & 0.164 & 0.473 & 380 & 0.155 & 0.110 & 380 \\
Ridge        & 1.591 & 1.654 & 0.753 & 1.654 & 125 & 0.518 & 0.462 & 125 \\
SVM          & 0.879 & 0.457 & 0.350 & 0.627 & 306 & 0.245 & 0.248 & 306 \\
\bottomrule
\end{tabular}
\end{table}

The irreducible-error floors $c$ tell the complementary story.
\autoref{fig:cvb} plots $(b,c)$ per fit. The two tree ensembles sit near the
desirable lower-left of the plot (modest $b$, low $c$). Boosting in
particular has a median floor of $\tilde c=0.208$ on classification (i.e.,
AUROC $\approx 0.79$) at full data---non-trivial but useful. Random Forest's
median $c=0.110$ is lower still. Linear/Logistic regression has a small floor
($\tilde c=0.139$) but only because its larger $b$ pulls error down; on
several classification cells it never overtakes Boosting in absolute error.
Lasso's floor is essentially $1$, reflecting the
shrink-to-mean failure mode discussed above. The takeaway is that for
practical decisions, $b$ alone is misleading; what matters is the joint
$(b,c)$.

\subsection{Approximate Shared Exponents Across Datasets}
\label{sec:universal}

The per-dataset exponents of \autoref{sec:exponents} cluster strongly by
model family. We test this directly: for each family we fit a single
shared exponent $b_{\text{model}}$ across all of its datasets while
leaving the per-dataset level parameters $(a_d, c_d)$ free
(\autoref{sec:crosscomb}). \autoref{tab:universal} reports the
comparison. Two facts must be read together. First,
$\Delta\mathrm{AIC}\in[+65,+908]$ \emph{strongly} favors the
unconstrained per-dataset fit on every family: the extra $D{-}1$ free
exponents capture residual structure that a single shared exponent
cannot. Second, despite this, the pooled $R^2$ gap is at most $0.011$
for 5 of 6 families. We therefore read the result as approximate
predictive compressibility rather than evidence for true equality of
dataset-level exponents: a shared exponent loses very little
explained variance on the pooled point cloud, while statistically real
per-dataset heterogeneity in $b$ remains.

\begin{table}[ht]
\caption{Shared-exponent vs.\ unconstrained joint fits per model family.
Shared $b$ with 95\% CI from the joint fit; $R^2_{\text{shared}}$ and
$R^2_{\text{unc}}$ are the joint-fit coefficients of determination of the
shared-exponent and unconstrained models on the same pooled point cloud;
$D$ datasets, $N$ pooled points. Lasso fails: its 95\% CI on $b$ crosses
zero and $R^2_{\text{shared}}=0.03$, consistent with the shrink-to-mean
pathology of \autoref{sec:fits}.}
\label{tab:universal}
\centering
\small
\begin{tabular}{lrrrrrrr}
\toprule
Model & shared $b$ & 95\% CI & $R^2_{\text{shared}}$ & $R^2_{\text{unc}}$ & $R^2$ gap & $D$ & $N$ \\
\midrule
Boosting     & 0.555 & [0.532, 0.579]  & 0.992 & 0.994 & 0.002 & 18 & 2{,}660 \\
RandomForest & 0.290 & [0.275, 0.306]  & 0.995 & 0.996 & 0.001 & 18 & 2{,}660 \\
SVM          & 0.709 & [0.661, 0.757]  & 0.970 & 0.980 & 0.011 & 18 & 2{,}142 \\
LinearModel  & 0.703 & [0.666, 0.741]  & 0.977 & 0.981 & 0.004 & 10 & 1{,}246 \\
Ridge        & 0.593 & [0.384, 0.803]  & 0.961 & 0.963 & 0.002 &  8 & 1{,}414 \\
\midrule
Lasso (fails)& 0.518 & [-2.19, 3.23]   & 0.032 & 0.033 & 0.000 &  8 & 1{,}414 \\
\bottomrule
\end{tabular}
\end{table}

\autoref{fig:universal_b_forest} visualises this with a forest plot:
the shared-exponent CI is tight for the four robust families
(Boosting, RandomForest, SVM, LinearModel) and overlaps with most of the
per-(dataset, model) pooled $b$'s as well as with the random-effects
meta-pooled estimate; Ridge is fragile and Lasso fails outright. A complementary
data-collapse analysis (rescaled $(\mathrm{error}-c)/a$ vs.\ $N$ in
log-log space) shows that the collapse is visibly partial---58\% of SVM
points and 32--46\% for the other families lie within $\pm 0.5$\,dex of
the shared-exponent line---consistent with real residual cell-level
heterogeneity, not perfect shared-exponent behaviour.

We stress-test this reading three ways (full detail in
\autoref{app:stress}). \textbf{(i)~LODO (leave-one-dataset-out)
transfer:} we refit the shared-$b$ model on $D{-}1$ datasets and use it to
predict the held-out dataset it was never fit on---a direct test of
whether the shared exponent \emph{transfers} to a new dataset rather than
merely describing the datasets it was trained on. This
gives median held-out $R^2$ gaps of $0.007$--$0.035$ for Boosting,
RandomForest, LinearModel, and SVM, with $b_{\text{shared}}^{(-d)}$
stable across folds (CV $0.04$--$0.09$); Ridge blows up
(CV $0.96$) and Lasso is meaningless. \textbf{(ii)~Bound sensitivity:}
relaxing $b\in[0,5]$ and doubling the $c$ bound leaves all five
non-Lasso shared exponents unchanged to within $0.001$. \textbf{(iii)~Permutation
null on the in-sample $R^2$ gap:} per-dataset exponent scrambling
($B{=}500$) yields one-sided $p\in[0.55,0.97]$, so the small in-sample
gap is not specific to a true shared exponent--- the held-out content
lives in the LODO transfer and the tight shared-$b$ CIs of
\autoref{tab:universal}, not the in-sample gap. The headline 5-of-6
count refines under LODO to: strong held-out support for Boosting,
RandomForest, LinearModel; moderate for SVM; fragile for Ridge;
failure for Lasso (which we treat as a negative control).

\subsection{Replicator-Implementation Variance}
\label{sec:variance}

\begin{figure}[ht]
\centering
\includegraphics[width=0.9\linewidth]{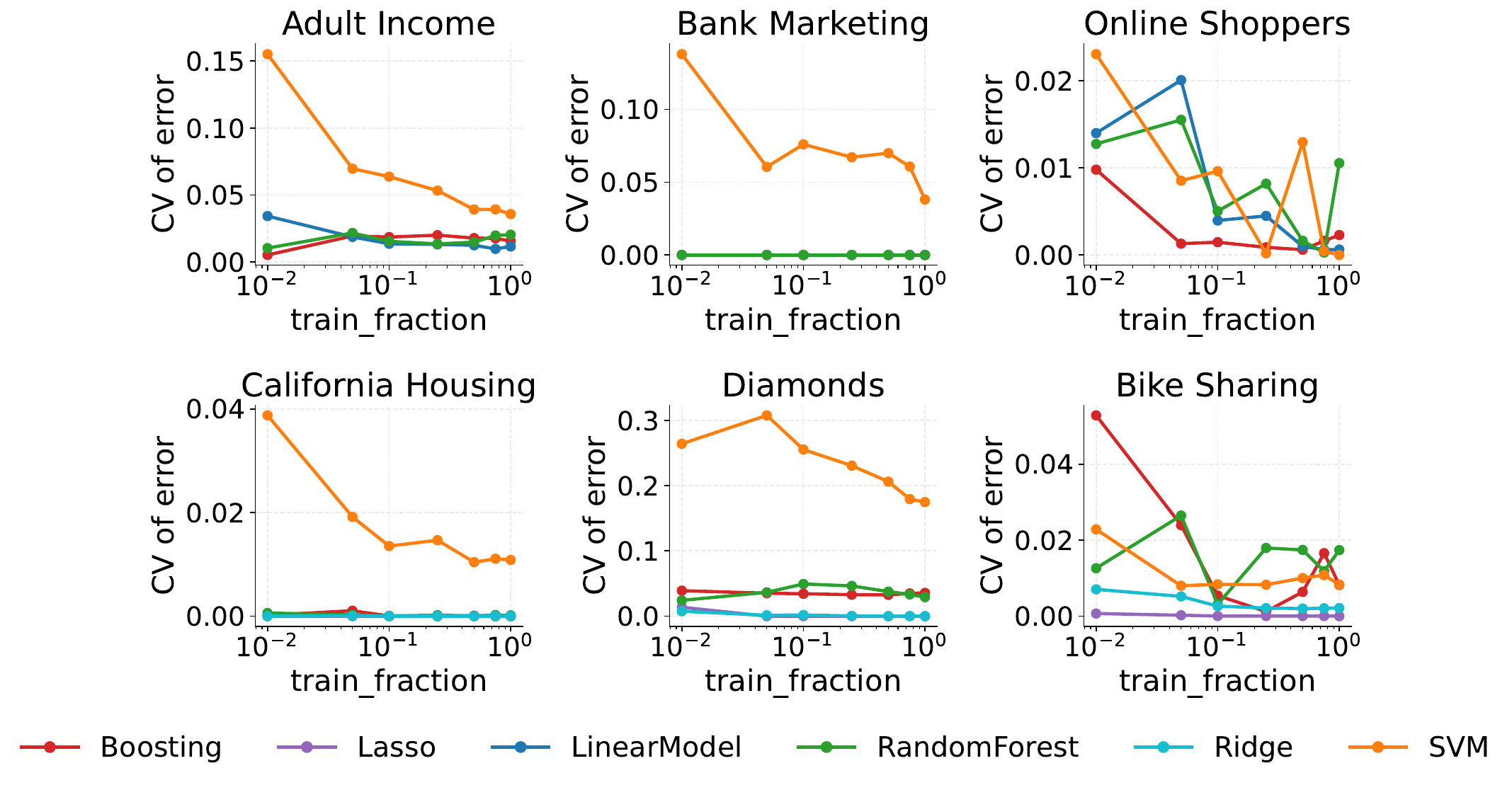}
\caption{Coefficient of variation of test error across replicators at each train
fraction, on representative datasets. Variance is largest at small $N$ and
shrinks but does not vanish at full data.}
\label{fig:cvar}
\end{figure}

The protocol fixes \texttt{random\_state=42}, so strictly compliant
re-implementation should yield byte-identical fits. Of 71 (dataset,
model) cells with $\ge 5$ replicators, 13 (18\%) have
$\mathrm{std}(b)<10^{-10}$, confirming the design floor. The remaining
58 cells show nonzero spread: unweighted mean
$\mathrm{CV}(b)=0.144$, size-weighted $0.131$, and $0.164$ excluding
the byte-identical cells (the upper bound on pure implementation
noise). With the seed fixed this is not seed/sampling variance: it
measures \emph{replicator-implementation} spread induced by the
unconstrained parts of the protocol (preprocessing, encoding,
missing-value handling). SVM dominates (per-family weighted CV $0.25$);
the other families cluster in $0.08$--$0.20$. By regime:
36 cells (51\%) at $\mathrm{CV}(b)<0.05$ (including the 13
byte-identical), 25 (35\%) at $0.05$--$0.30$ (legitimate protocol
ambiguity), and 10 (14\%) at $\ge 0.30$ (more consistent with
implementation bugs; e.g.\ RandomForest$\times$\texttt{diabetes\_pima}
$=0.72$). \autoref{fig:cvar} shows this dispersion grows at small $N$
where preprocessing choices have the largest leverage.

\begin{figure}[ht]
\centering
\includegraphics[width=\linewidth]{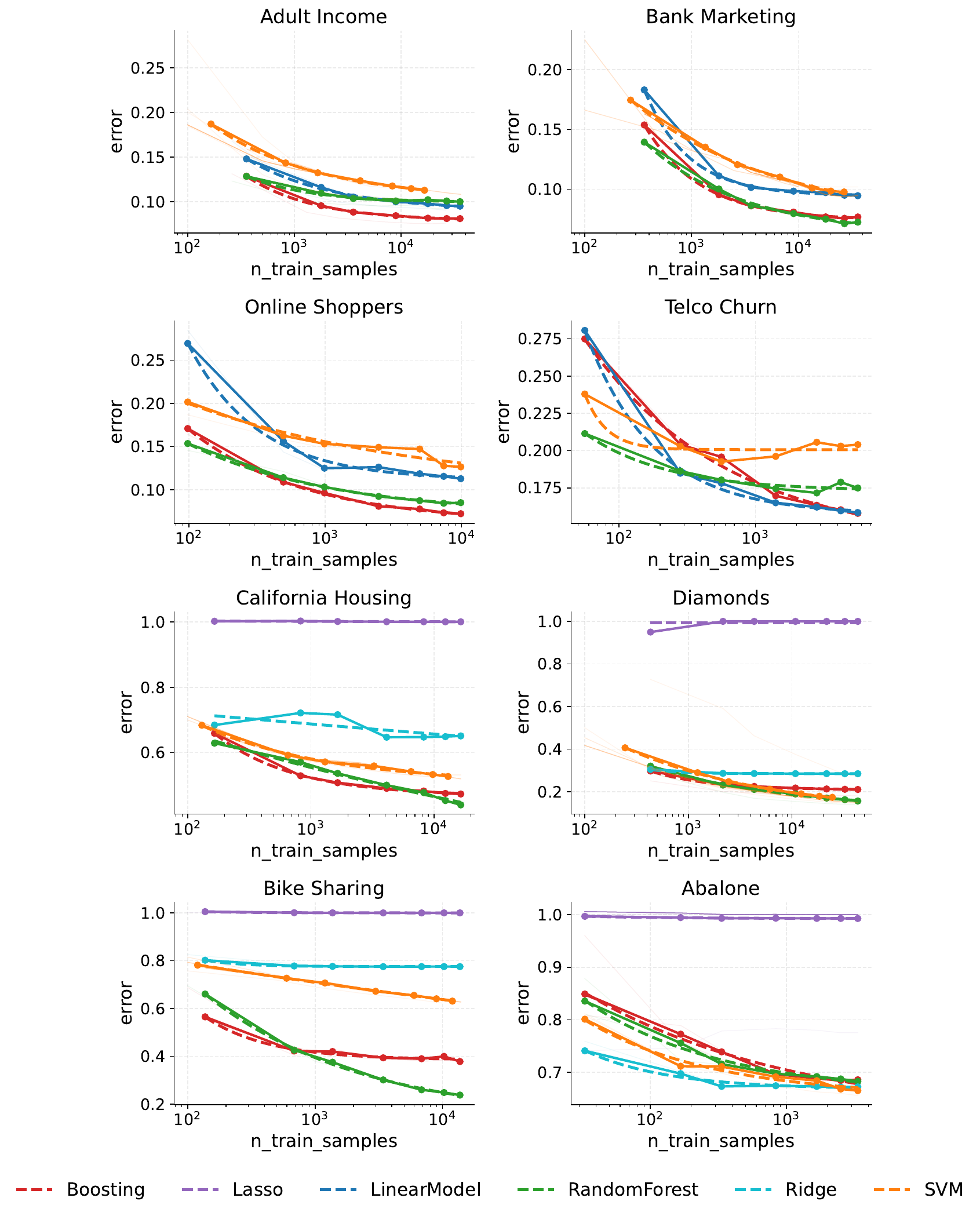}
\caption{Pooled scaling curves on 4 representative classification (top row)
and 4 representative regression (bottom row) datasets. Thin lines: per-replicator
curves; thick line: across-replicator mean; dashed line: pooled power-law fit.
Per-replicator dispersion at small $N$ is visible; pooled fits are visibly
tighter than any single replicator's curve.}
\label{fig:grid}
\end{figure}

The practical implication: single-group scaling-law studies should
report uncertainty intervals on $b$ wide enough to accommodate this
14\% floor; comparative claims (``model A has larger $b$ than B'')
should require at minimum a $\sim 0.15\,b$ gap to be credible against
protocol-drift noise. Pooling across replicators
(\autoref{fig:grid}, dashed) gives tighter fits than any single
replicator's curve and underlies the data-requirement table below.

\subsection{Data Requirements and Model Ranking}
\label{sec:datareq}

From the pooled per-cell power-law fit, we invert \autoref{eq:powerlaw} to
get the sample size $N^\star$ required to reach a target error of $0.15$.
\autoref{tab:datareq} reports $N^\star$ for each classification (dataset,
model) cell where the target is reachable (i.e., where $c<0.15$). On easy
classification problems---\texttt{adult\_income}, \texttt{online\_shoppers}---tree
ensembles get to $1-\mathrm{AUROC}=0.15$ with fewer than 400 training
points. At the other extreme, \texttt{telco\_churn} needs $\sim 16{,}500$
points for Boosting to reach the target.

We flag $N^\star$ values requiring $>10\times$ extrapolation beyond the
largest observed $N$ with a dagger ($\dagger$); the extreme case is
\texttt{german\_credit}$\times$LinearModel ($b=0.038$, $N^\star\approx
2.8\times10^9$), which should read as ``unattainable under linear
scaling.'' Tiny $N^\star$ values are flagged ``$<30$''
(degenerate fits with $a$ small and $c{\approx}0$). Cells with
$c\ge 0.15$ are unreachable without changing the model class.
\autoref{fig:dr} visualizes the full pattern.

\begin{table}[ht]
\caption{Pooled-fit sample size $N^\star$ to reach target error $0.15$
on representative classification cells where the floor $c<0.15$. Full
table (including pathological extrapolations and degenerate fits) in
\autoref{tab:datareq_full}.}
\label{tab:datareq}
\centering
\small
\begin{tabular}{llrrrr}
\toprule
Dataset & Model & $a$ & $b$ & $c$ & $N^\star$ \\
\midrule
adult\_income       & Boosting     &  2.669 & 0.679 & 0.079 &        208 \\
adult\_income       & RandomForest &  1.723 & 0.695 & 0.099 &        159 \\
bank\_marketing     & Boosting     &  9.940 & 0.820 & 0.074 &        383 \\
bank\_marketing     & RandomForest &  1.246 & 0.473 & 0.063 &        276 \\
online\_shoppers    & Boosting     &  1.163 & 0.517 & 0.062 &        148 \\
online\_shoppers    & RandomForest &  0.553 & 0.416 & 0.071 &        109 \\
online\_shoppers    & LinearModel  &  7.618 & 0.844 & 0.111 &        518 \\
telco\_churn        & Boosting     &  0.737 & 0.417 & 0.137 & 16{,}539 \\
wine\_quality       & RandomForest &  0.528 & 0.194 & 0.000 &        666 \\
\bottomrule
\end{tabular}
\end{table}

Finally, \autoref{fig:rank} ranks the six families at full data:
Boosting wins 50\% of datasets (avg.\ rank $1.78$), RandomForest 33\%
($2.11$), LinearModel and SVM mid-pack, Ridge and Lasso $0\%$
($3.75$, $5.00$). The ``boosted trees rule tabular'' result reproduces
cleanly. With only 17--30 replicators per cell and small test sets on the
smallest datasets, win-rates should be read as effect-size summaries,
not significance tests.

\section{Discussion}
\label{sec:discussion}

\paragraph{For practitioners.}
Default-tuned tree ensembles dominate at full data, reproducing
\citet{grinsztajn2022why} on a broader list and adding that they also
have the most favorable $(b,c)$ profile. The data-requirement table
gives concrete estimates: $\sim$200--500 rows suffice to push
$1-\mathrm{AUROC}<0.15$ on \texttt{adult\_income},
\texttt{bank\_marketing}, and \texttt{online\_shoppers} \citep{dua2019uci} with a booster,
while \texttt{telco\_churn} needs an order of magnitude more. Several
cells have $c$ above the practical target---more data cannot help, and
the right move is a better feature set, loss, or model class.

\paragraph{For theory.}
\citet{kaplan2020scaling} report $b\approx 0.095$ for transformer loss
and \citet{hestness2017deep} $b\approx 0.07$--$0.35$ across modalities.
Our classical exponents (median $0.495$; $0.32$--$0.48$ for tree
ensembles; $0.83$ for LinearModel on classification) are if anything
\emph{larger} than the small-NN exponents---each row buys more error
reduction in classical tabular models than each token in a frontier
transformer. What distinguishes the classical regime is the
irreducible-error floor: tree ensembles routinely have $c\approx
0.10$--$0.21$ on classification, orders of magnitude larger than
overparameterised LMs. In tabular ML the floor, not the exponent, is
the binding constraint.

\paragraph{Methodology.}
Treating a graduate cohort of 127 as a population of independent
replicators gives a near-free estimate of the protocol-drift floor on
a scaling-law claim; the $\mathrm{CV}(b)=0.144$ figure is a concrete
prior for single-group studies. Pooling across replicators
(\autoref{fig:grid}) also yields tighter fits than any individual
curve at no extra compute cost.

\section{Limitations}
\label{sec:limitations}

The scope conditions below are deliberate design choices, not oversights.
Each keeps a distributed, 127-replicator study tractable and its
shared-exponent and reproducibility comparisons clean; we state them as the
conditions under which our claims hold, not as afterthoughts.

(i)~\textbf{Fixed hyperparameters} (no per-$N$ tuning), so our $b$
values are lower bounds for what proper model selection could achieve;
Lasso/Ridge are most fragile under this constraint. We fix them
deliberately: a single shared configuration is what makes 127 independent
re-implementations directly comparable and isolates the data-size effect
from tuning noise.
(ii)~\textbf{Dataset coverage}: 18 small-to-medium clean datasets
(largest 53{,}940 rows); we say nothing about million-row settings.
(iii)~\textbf{SVM cap}: classification SVMs truncated to 10{,}000-row
subsample, slightly suppressing large-$N$ behavior.
(iv)~\textbf{Single split seed}: with \texttt{random\_state=42} the
$\mathrm{CV}(b)=0.144$ is replicator-implementation spread, not seed
variance; a multi-seed study would report a larger total CV.
(v)~\textbf{Imperfect replicators and partial blocking}: each replicator
saw only 3/18 datasets, so per-cell CV is computed over different
sub-cohorts; cells with $\mathrm{CV}(b)\ge 0.30$ (10/71) are more
consistent with implementation bugs than legitimate ambiguity and we
treat them as part of the phenomenon being measured.
(vi)~\textbf{Shared-exponent stress tests are descriptive}: supported
in-sample (small gaps) and held-out (LODO for four families) but
in-sample gap lacks specificity under permutation null and Ridge is
fragile across folds (\autoref{app:stress}).
(vii)~\textbf{Agreement is not correctness}: our shared-exponent
evidence rests on independent per-replicator fits agreeing (low
cross-replicator $\mathrm{CV}(b)$), but agreement among independent
estimators is necessary, not sufficient---shared implementation
conventions can manufacture consensus---so low cross-replicator variance
corroborates rather than proves a universal exponent
\citep{ding2026llmagree}.

\section{Conclusion}
\label{sec:conclusion}

A 127-replicator benchmark study (18 datasets, 6 families) shows $aN^{-b}+c$
fits 77.7\% of cells at $R^2{>}0.8$; a family-level exponent
compresses cross-dataset curves for 5/6 families (Lasso fails);
tree ensembles dominate at full data; fixed-seed CV$(b)\!\approx\!0.14$.

\begin{ack}
I thank the graduate-course cohort whose submitted model-output runs constitute the
dataset analyzed here. All analysis, framing, and conclusions are the author's own.

The author acknowledges the National Artificial Intelligence Research Resource (NAIRR)
Pilot and the Jetstream2 cloud resource at Indiana University for contributing to this
research result. This work used Jetstream2 through NAIRR Pilot allocation NAIRR250223;
Jetstream2 is supported by the U.S. National Science Foundation under award NSF-OAC
2005506. The author also thanks the Wharton AI \& Analytics Initiative at the University
of Pennsylvania for financially supporting this project through its AI Education
Innovation Fund.

\end{ack}

\bibliographystyle{plainnat}
\bibliography{references}

\appendix

\section{Stress Tests on the Shared-Exponent Claim}
\label{app:stress}

This appendix expands the three stress tests summarised in
\autoref{sec:universal}.

\paragraph{Caveats on reading the in-sample $R^2$ gap.} Several caveats
sharpen the in-sample reading of \autoref{tab:universal}. First, the
joint model retains $2D$ free per-dataset $(a_d,c_d)$ parameters
(36 nuisance parameters for the 18-dataset families), so a small
pooled $R^2$ gap is necessary but not sufficient evidence for
exponent equality; we do not interpret it as such. Second, the $R^2$
gap is itself a weak goodness-of-fit summary for nonlinear least
squares with free dataset offsets; we report it because it is
directly interpretable (an $R^2$ gap of $0.011$ corresponds to roughly
1\% of explained-variance loss), not as a hypothesis test. Third,
Ridge's CI is wide ($[0.38, 0.80]$) because two of its eight cells
hit the upper-bound $b{=}3$ in their per-cell fits; we include Ridge
in the 5/6 count for completeness but treat its shared-exponent
estimate as fragile. Fourth, Lasso fails outright (CI crosses zero,
$R^2_{\text{shared}}=0.03$): we treat it as a \emph{negative control}
demonstrating that a small gap alone is insufficient evidence for
shared-exponent structure.

\paragraph{Held-out evidence: leave-one-dataset-out (LODO) shared-$b$
prediction.} To probe whether the small in-sample $R^2$ gap reflects a
genuinely transferable exponent or merely an in-sample compression
artefact of the $2D$ free $(a_d,c_d)$ offsets, we re-fit the shared-$b$
joint model on each family's $D{-}1$ remaining datasets, freeze
$b_{\text{shared}}^{(-d)}$, and on the held-out dataset $d$ fit only
$(a_d, c_d)$; we then compare against the unconstrained 3-parameter fit
on the same held-out points (\texttt{lodo\_shared\_b.parquet};
re-run via \texttt{analysis/scripts/lodo\_shared\_b.py}). Median
held-out $R^2$ gaps are $0.007$ (Boosting), $0.010$ (RandomForest),
$0.035$ (SVM), $0.012$ (LinearModel), $0.088$ (Ridge), $0.0002$ (Lasso),
with a fraction of folds within a $0.05$ gap of $83.3\%$, $83.3\%$,
$61.1\%$, $80.0\%$, $12.5\%$, and $75.0\%$ respectively. The held-out
$b_{\text{shared}}^{(-d)}$ values are stable for the four well-behaved
families (CV across folds $0.036$--$0.085$) and unstable for Ridge
(CV $=0.96$, with one fold landing at $b{=}2.94$ when
\texttt{insurance\_charges} is held out). Across Boosting, RandomForest,
and LinearModel the held-out gap is within roughly an order of magnitude
of the in-sample gap and 80\%+ of folds transfer cleanly, supporting
the predictive-compressibility reading. SVM's held-out gap is
$\sim 3{\times}$ its in-sample value (honest mild inflation), Ridge's
LODO gap blows up by $\sim 40{\times}$ confirming fragility, and Lasso
is meaningless under either evaluation.

\paragraph{Bound sensitivity.} Refitting the joint model with relaxed
bounds ($b\in[0,5]$, $c$ doubled) leaves all five non-Lasso shared
exponents unchanged to within $0.001$ and Ridge's LODO
$\mathrm{CV}(b_{\text{loo}})=0.96$ unchanged to three decimal places.
Lasso's $b$ drifts to whichever ceiling is imposed
($R^2\approx 0.03$ either way), confirming its failure is intrinsic.

\paragraph{Permutation null on the in-sample $R^2$ gap.} Because the
joint shared-$b$ model retains $2D$ free per-dataset $(a_d,c_d)$
offsets, a small in-sample $R^2$ gap could be a flexibility artefact of
the 36-parameter offset bath rather than evidence of a real common
exponent. We test this with a per-dataset exponent-scrambling
permutation (\texttt{analysis/scripts/permutation\_null\_shared\_b.py},
$B{=}500$, \texttt{seed=42}): each iteration permutes $\hat b_d$ across
datasets within a family while holding $(\hat a_d, \hat c_d)$ fixed,
simulates $\text{error}^{\text{null}}_{d,i}=\hat a_d
N^{-b_{\pi(d)}}+\hat c_d+\epsilon_{d,i}$ with Gaussian $\epsilon$ at the
per-cell empirical residual std, and refits the shared-$b$ joint model
plus its per-cell unconstrained reference on the null panel. The
observed in-sample gaps are \emph{not} unusually small under this null:
empirical one-sided $p$-values (Variant~B, $\Pr(\Delta R^2_{\text{null}}
\le \Delta R^2_{\text{obs}})$) are $0.59$ (Boosting), $0.97$
(RandomForest), $0.85$ (SVM), $0.55$ (LinearModel), $0.77$ (Ridge), and
$0.94$ (Lasso), with null-gap medians $0.0012$, $0.0005$, $0.0021$,
$0.0022$, $0.0001$, and $0.0000$. A cheaper Variant~A that shuffles
dataset labels among normalised $(\text{error}-\hat c_d)/\hat a_d$
curves gives $p\in[0.54,0.91]$. We therefore use the in-sample gap
only as a compression metric, not as evidence for true
dataset-independent exponents. Per-family null distributions are
stored under \texttt{permutation\_null\_shared\_b} in
\texttt{results\_summary.json}.

\section{Full Data-Requirement Table}
\label{app:datareq}

\begin{table}[ht]
\caption{Full pooled-fit sample size $N^\star$ to reach target error
$0.15$ on classification cells where the floor $c<0.15$.
$\dagger$~marks $N^\star$ requiring extrapolation more than $10\times$
beyond the largest observed $N$ for that dataset; these values are
statistically unreliable and shown for completeness only. ``$<30$''
marks cells whose fit predicts the target is already met below the
smallest training-fraction bin we actually evaluated, indicating a
degenerate fit (small $a$, $c{\approx}0$) rather than a genuine claim
that fewer than 30 rows suffice.}
\label{tab:datareq_full}
\centering
\small
\begin{tabular}{llrrrr}
\toprule
Dataset & Model & $a$ & $b$ & $c$ & $N^\star$ \\
\midrule
adult\_income       & Boosting     &  2.669 & 0.679 & 0.079 &        208 \\
adult\_income       & LinearModel  &  1.504 & 0.556 & 0.091 &        338 \\
adult\_income       & RandomForest &  1.723 & 0.695 & 0.099 &        159 \\
adult\_income       & SVM          &  0.708 & 0.438 & 0.102 &        474 \\
bank\_marketing     & Boosting     &  9.940 & 0.820 & 0.074 &        383 \\
bank\_marketing     & LinearModel  & 43.044 & 1.051 & 0.095 &        564 \\
bank\_marketing     & RandomForest &  1.246 & 0.473 & 0.063 &        276 \\
bank\_marketing     & SVM          &  0.463 & 0.274 & 0.067 &        515 \\
diabetes\_pima      & Boosting     &  0.692 & 0.213 & 0.000 &      1{,}315 \\
diabetes\_pima      & RandomForest &  0.472 & 0.147 & 0.000 &      2{,}425 \\
diabetes\_pima      & SVM          &  0.707 & 0.195 & 0.000 &      2{,}831 \\
employee\_attrition & LinearModel  &  3.392 & 0.647 & 0.139 &      7{,}364 \\
employee\_attrition & SVM          &  2.099 & 0.404 & 0.043 &      1{,}589 \\
german\_credit      & LinearModel  &  0.343 & 0.038 & 0.000 & $2.8{\times}10^{9}\,\dagger$ \\
german\_credit      & RandomForest &  0.396 & 0.229 & 0.131 & $556{,}439\,\dagger$ \\
heart\_disease      & Boosting     & 17.120 & 1.409 & 0.039 &         36 \\
heart\_disease      & LinearModel  &  0.735 & 0.454 & 0.000 &         33 \\
heart\_disease      & RandomForest &  0.126 & 0.161 & 0.000 &     $<30$ \\
heart\_disease      & SVM          &  0.379 & 0.504 & 0.028 &     $<30$ \\
online\_shoppers    & Boosting     &  1.163 & 0.517 & 0.062 &        148 \\
online\_shoppers    & LinearModel  &  7.618 & 0.844 & 0.111 &        518 \\
online\_shoppers    & RandomForest &  0.553 & 0.416 & 0.071 &        109 \\
online\_shoppers    & SVM          &  0.320 & 0.251 & 0.099 &      1{,}504 \\
telco\_churn        & Boosting     &  0.737 & 0.417 & 0.137 & 16{,}539 \\
wine\_quality       & Boosting     &  0.629 & 0.277 & 0.079 &  2{,}670 \\
wine\_quality       & RandomForest &  0.528 & 0.194 & 0.000 &        666 \\
\bottomrule
\end{tabular}
\end{table}

\section{Additional Figures}
\label{app:figures}

This appendix collects supporting figures referenced from the main text.

\begin{figure}[ht]
\centering
\includegraphics[width=0.7\linewidth]{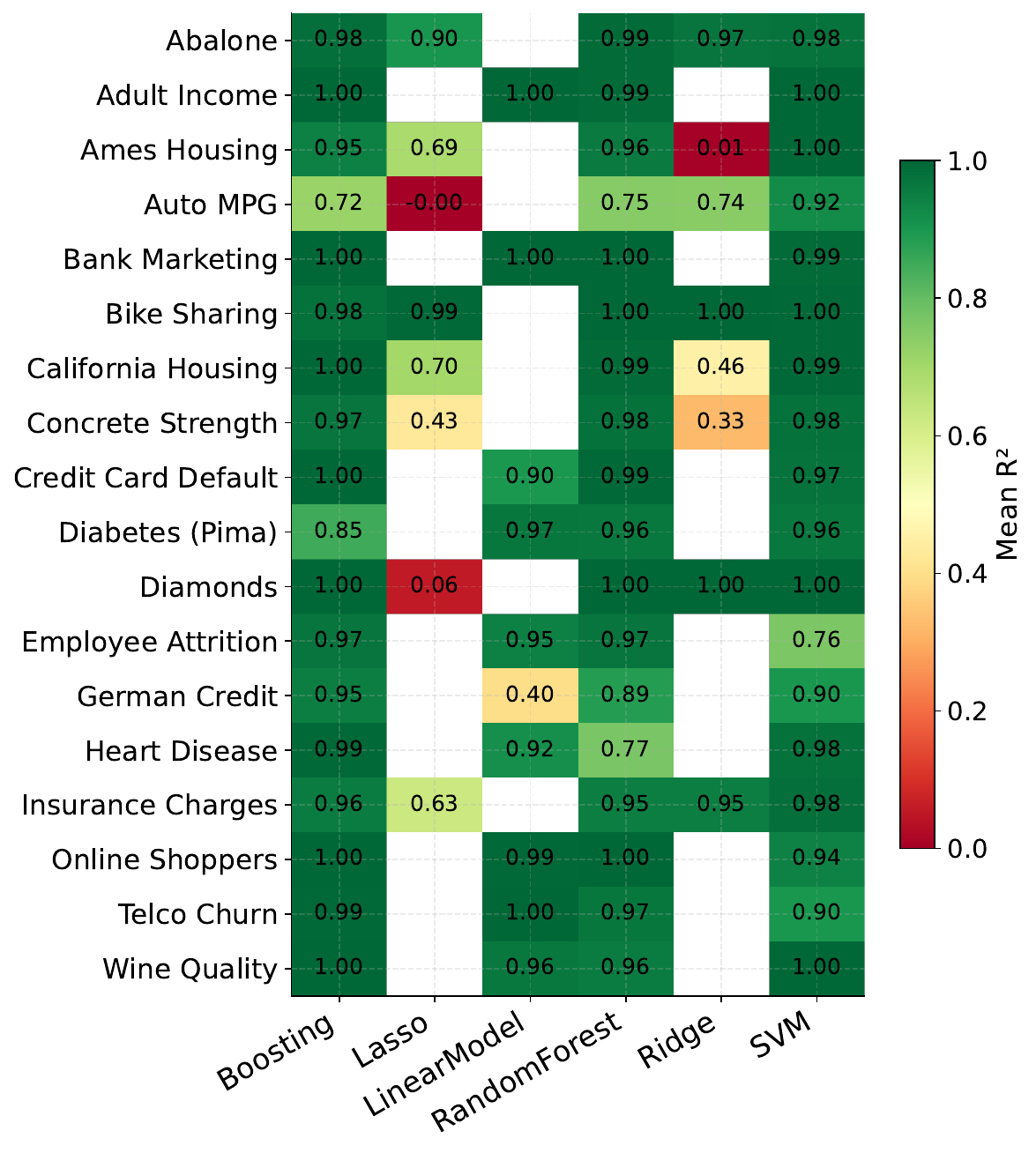}
\caption{Mean per-replicator $R^2$ of power-law fits per (dataset, model) cell.
Most cells exceed 0.9; diagnostic dark cells (Lasso, Ridge, and a notable
Lasso/Ridge column on \texttt{california\_housing}) flag cases where the
power-law assumption is misleading. Companion to \autoref{tab:r2}.}
\label{fig:r2}
\end{figure}

\begin{figure}[ht]
\centering
\includegraphics[width=0.7\linewidth]{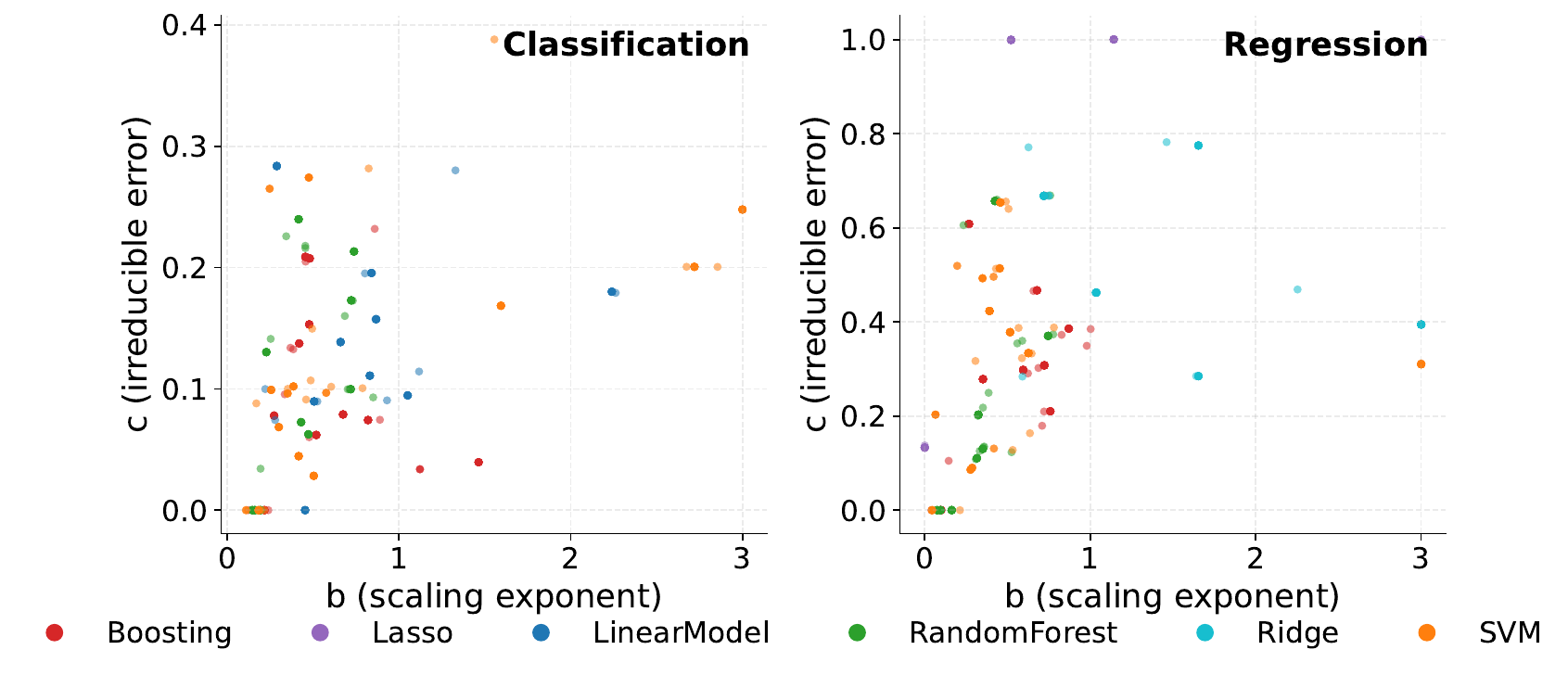}
\caption{Per-(replicator, dataset, model) scatter of $(b,c)$, faceted by task.
Tree ensembles occupy the favorable low-$c$, moderate-$b$ region; Lasso
clusters at $c\approx1$ on classification.}
\label{fig:cvb}
\end{figure}

\begin{figure}[ht]
\centering
\includegraphics[width=0.7\linewidth]{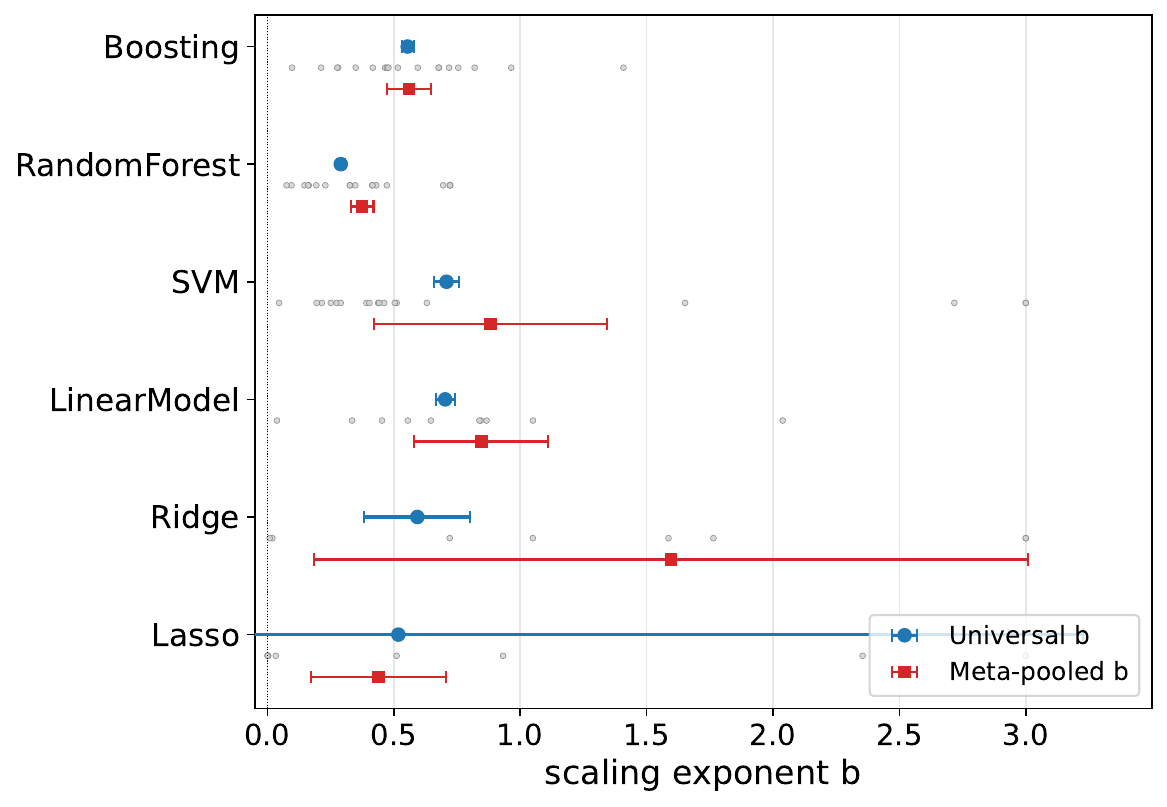}
\caption{Forest plot of scaling exponents per model family. Square: the
shared $b$ from the shared-exponent joint fit (\autoref{tab:universal})
with 95\% CI. Circles: per-(dataset, model) pooled $b$'s. Diamond: the
DerSimonian--Laird random-effects meta-pooled $b$ across cells. Note the
four tight shared-$b$ intervals for the robust families (Boosting,
RandomForest, SVM, LinearModel), Ridge's wider/fragile CI, and Lasso's
CI that crosses zero.}
\label{fig:universal_b_forest}
\end{figure}

\begin{figure}[ht]
\centering
\begin{minipage}[t]{0.48\linewidth}
\centering
\includegraphics[width=\linewidth]{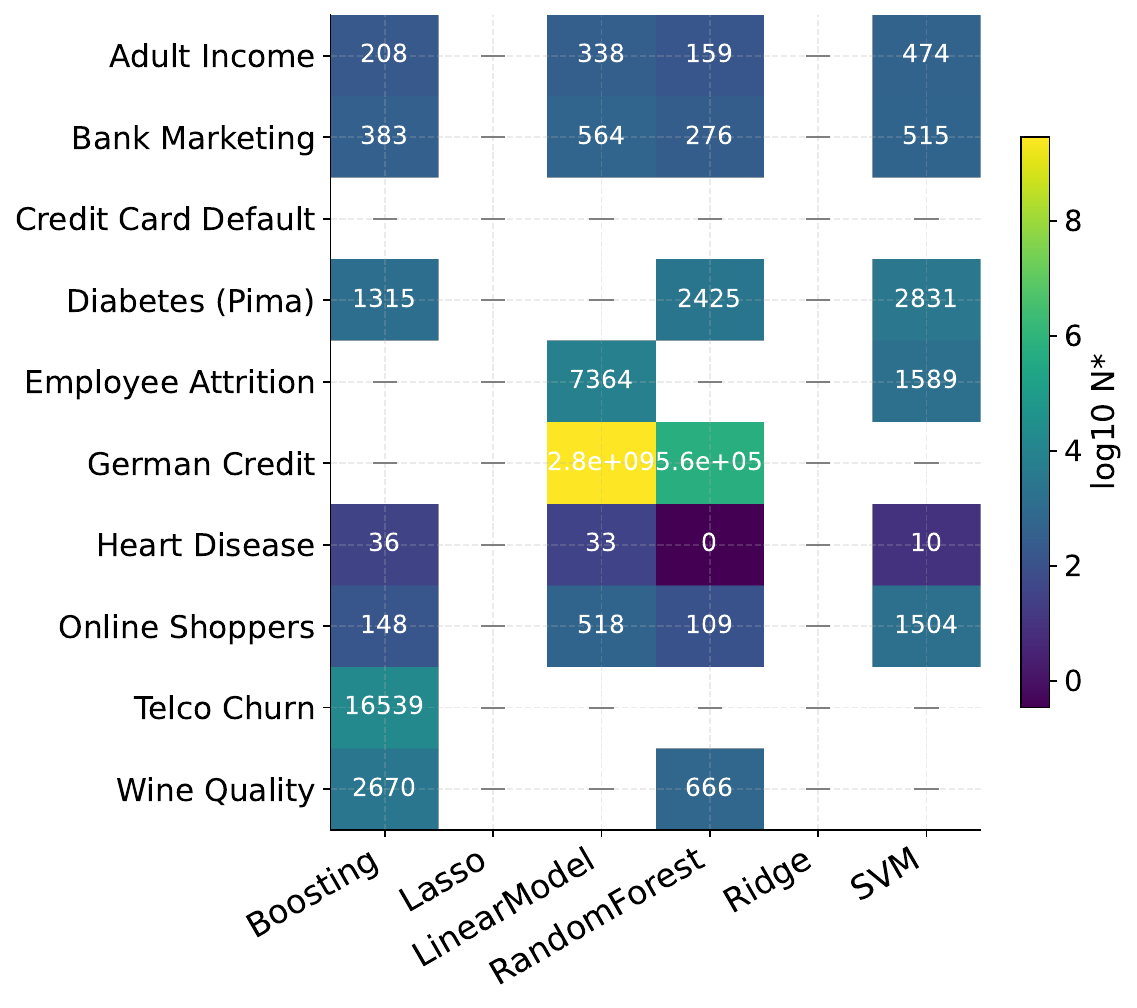}
\caption{$\log_{10} N^\star$ to reach error $0.15$ on each (classification
dataset, model) cell. Empty cells indicate $c\ge 0.15$ (unreachable).
Companion to \autoref{tab:datareq}.}
\label{fig:dr}
\end{minipage}\hfill
\begin{minipage}[t]{0.48\linewidth}
\centering
\includegraphics[width=\linewidth]{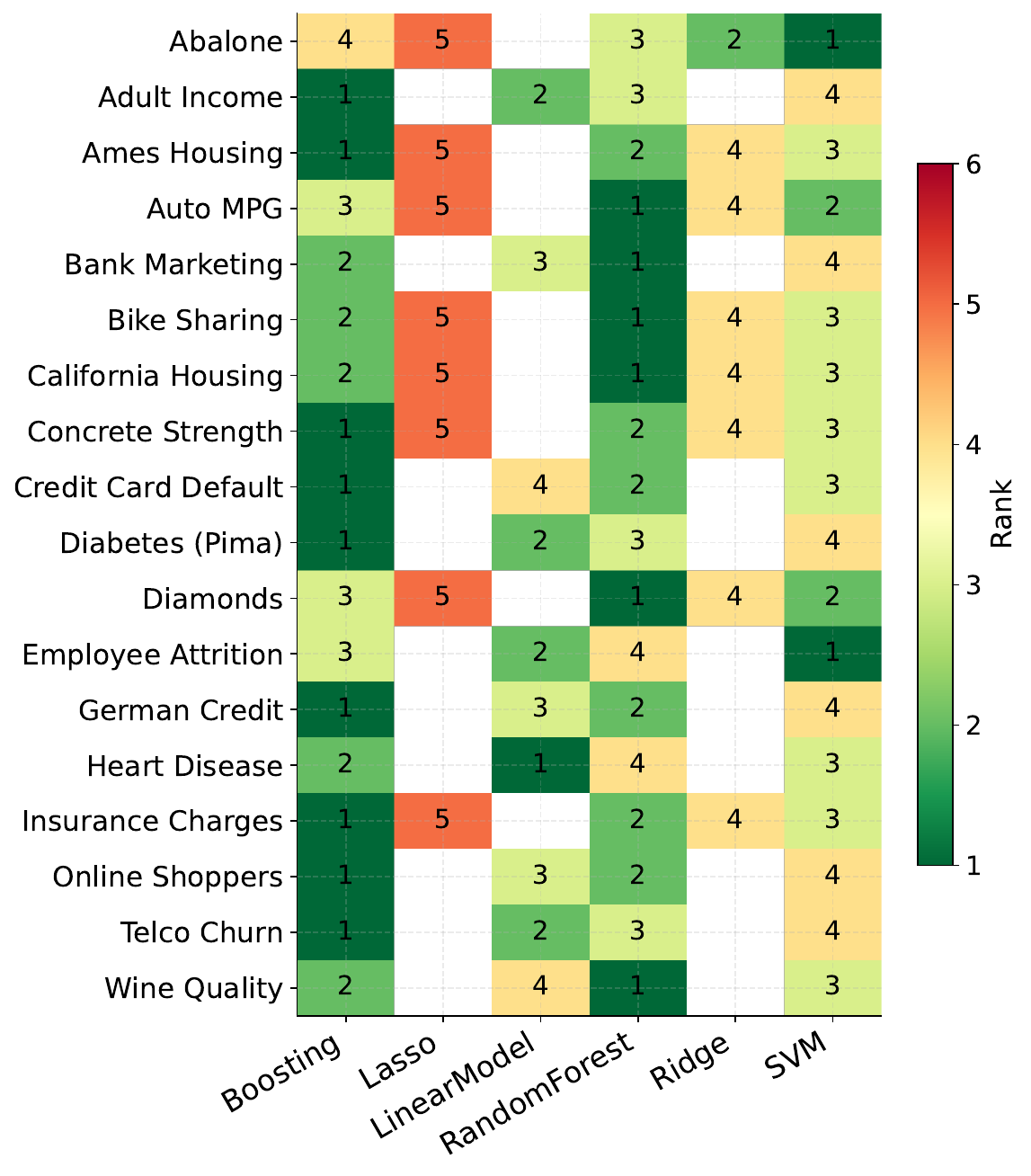}
\caption{Per-dataset model rank at training fraction $1.0$ (1 = best).
Boosting wins 50\% of datasets; tree ensembles together win 83\%.}
\label{fig:rank}
\end{minipage}
\end{figure}

\newpage
\section*{NeurIPS Paper Checklist}

\begin{enumerate}

\item {\bf Claims}
    \item[] Question: Do the main claims made in the abstract and introduction accurately reflect the paper's contributions and scope?
    \item[] Answer: \answerYes{}
    \item[] Justification: The abstract and introduction enumerate three concrete findings (power-law fits, approximate shared exponents, replicator-implementation variance) that map directly to \autoref{sec:fits}, \autoref{sec:universal}, and \autoref{sec:variance}.

\item {\bf Limitations}
    \item[] Question: Does the paper discuss the limitations of the work performed by the authors?
    \item[] Answer: \answerYes{}
    \item[] Justification: \autoref{sec:limitations} enumerates six limitations (fixed hyperparameters, dataset coverage, SVM cap, single split seed, imperfect replicators / partial blocking, descriptive nature of shared-exponent stress tests).

\item {\bf Theory Assumptions and Proofs}
    \item[] Question: For each theoretical result, does the paper provide the full set of assumptions and a complete (and correct) proof?
    \item[] Answer: \answerNA{}
    \item[] Justification: The paper is empirical; no theoretical results are claimed.

\item {\bf Experimental Result Reproducibility}
    \item[] Question: Does the paper fully disclose all the information needed to reproduce the main experimental results of the paper to the extent that it affects the main claims and/or conclusions of the paper (regardless of whether the code and data are provided or not)?
    \item[] Answer: \answerYes{}
    \item[] Justification: \autoref{sec:methodology} specifies the protocol (datasets, model families, training fractions, fixed seed, fitting bounds), and we will release aggregated $(N,\mathrm{error})$ CSVs, per-cell pooled fits, and analysis scripts upon acceptance.

\item {\bf Open access to data and code}
    \item[] Question: Does the paper provide open access to the data and code, with sufficient instructions to faithfully reproduce the main experimental results, as described in supplemental material?
    \item[] Answer: \answerYes{}
    \item[] Justification: Aggregated curves, per-cell fits, the data-requirement table, and analysis scripts will be released at an anonymous mirror during review (\autoref{sec:crosscomb}); all 18 datasets used are public.

\item {\bf Experimental Setting/Details}
    \item[] Question: Does the paper specify all the training and test details (e.g., data splits, hyperparameters, how they were chosen, type of optimizer, etc.) necessary to understand the results?
    \item[] Answer: \answerYes{}
    \item[] Justification: \autoref{sec:protocol} specifies the 80/20 split with seed 42, the seven nested training fractions, default-style hyperparameters held fixed across $N$, and the SVM cap; \autoref{sec:fitting} specifies the power-law fit bounds and $R^2$ thresholds.

\item {\bf Experiment Statistical Significance}
    \item[] Question: Does the paper report error bars suitably and correctly defined or other appropriate information about the statistical significance of the experiments?
    \item[] Answer: \answerYes{}
    \item[] Justification: \autoref{tab:universal} reports 95\% CIs on shared exponents from joint nonlinear least-squares; \autoref{sec:variance} reports $\mathrm{CV}(b)$ across replicators with weighted and unweighted variants; \autoref{app:stress} reports LODO held-out gaps and a permutation null with 500 replicates.

\item {\bf Experiments Compute Resources}
    \item[] Question: For each experiment, does the paper provide sufficient information on the computer resources (type of compute workers, memory, time of execution) needed to reproduce the experiments?
    \item[] Answer: \answerNo{}
    \item[] Justification: All training runs use \texttt{scikit-learn} with default-style hyperparameters on commodity hardware; per-run cost is small (seconds to minutes) and the experiments were distributed across the 127 replicators' personal laptops, so we did not aggregate compute hours. Aggregated analysis scripts run in minutes on a single laptop.

\item {\bf Code Of Ethics}
    \item[] Question: Does the research conducted in the paper conform, in every respect, with the NeurIPS Code of Ethics?
    \item[] Answer: \answerYes{}
    \item[] Justification: The work uses only public benchmark datasets, fits descriptive statistical models, and does not deploy systems or release models that could enable misuse.

\item {\bf Broader Impacts}
    \item[] Question: Does the paper discuss both potential positive societal impacts and negative societal impacts of the work performed?
    \item[] Answer: \answerNA{}
    \item[] Justification: The paper is foundational empirical methodology (scaling-law fits and replicator-variance estimates for tabular ML); it does not introduce new applications or models with foreseeable societal impacts beyond standard ML benchmarking practice.

\item {\bf Safeguards}
    \item[] Question: Does the paper describe safeguards that have been put in place for responsible release of data or models that have a high risk for misuse (e.g., pretrained language models, image generators, or scraped datasets)?
    \item[] Answer: \answerNA{}
    \item[] Justification: The released artifacts are aggregated learning curves and pooled power-law fits on public tabular datasets; they pose no misuse risk.

\item {\bf Licenses for existing assets}
    \item[] Question: Are the creators or original owners of assets (e.g., code, data, models), used in the paper, properly credited and are the license and terms of use explicitly mentioned and properly respected?
    \item[] Answer: \answerYes{}
    \item[] Justification: All 18 datasets are public and cited (\autoref{tab:datasets}); model implementations come from \texttt{scikit-learn} (BSD-3-Clause). Our released analysis code uses a permissive open-source license.

\item {\bf New Assets}
    \item[] Question: Are new assets introduced in the paper well documented and is the documentation provided alongside the assets?
    \item[] Answer: \answerYes{}
    \item[] Justification: The aggregated learning-curve CSVs, per-cell pooled fits, data-requirement table, and analysis scripts are released with documentation describing column schemas, fitting procedures, and reproduction commands.

\item {\bf Crowdsourcing and Research with Human Subjects}
    \item[] Question: For crowdsourcing experiments and research with human subjects, does the paper include the full text of instructions given to participants and screenshots, if applicable, as well as details about compensation (if any)?
    \item[] Answer: \answerNA{}
    \item[] Justification: This was not crowdsourcing or human-subjects research. The 127 replicators were members of a graduate ML cohort and submitted course assignments; the paper analyses the resulting model-fit artifacts, not the individuals themselves. The protocol skeleton (the code template every cohort member received) is included in the release.

\item {\bf Institutional Review Board (IRB) Approvals or Equivalent for Research with Human Subjects}
    \item[] Question: Does the paper describe potential risks incurred by study participants, whether such risks were disclosed to the subjects, and whether Institutional Review Board (IRB) approvals (or an equivalent approval/review based on the requirements of your country or institution) were obtained?
    \item[] Answer: \answerNA{}
    \item[] Justification: As noted above, this is not human-subjects research; only de-identified model-fit outputs from a course assignment are analysed and only in aggregate.

\item {\bf Declaration of LLM usage}
    \item[] Question: Does the paper describe the usage of LLMs if it is an important, original, or non-standard component of the core methods in this research?
    \item[] Answer: \answerNA{}
    \item[] Justification: LLMs were not used as any component of the core methodology. The core methods (power-law fitting, joint shared-exponent estimation, LODO transfer, permutation null) are all classical numerical methods implemented in \texttt{scipy} / \texttt{scikit-learn}. Any LLM use was limited to writing, editing, and formatting and is not part of the research method.

\end{enumerate}

\end{document}